\pdfoutput=1

\documentclass[11pt]{article}

\usepackage[]{EMNLP2023}

\usepackage{times}
\usepackage{latexsym}
\usepackage{graphicx}

\usepackage[T1]{fontenc}

\usepackage[utf8]{inputenc}

\usepackage{microtype}

\usepackage{inconsolata}

\usepackage[utf8]{inputenc}
\usepackage{enumitem}
\usepackage{blindtext}
\usepackage{float}
\title{Don't Waste a Single Annotation: Improving Single-Label Classifiers Through Soft Labels}

\author{Ben Wu, Yue Li, Yida Mu, Carolina Scarton, Kalina Bontcheva \and Xingyi Song \\
    Department of Computer Science, The University of Sheffield, Sheffield, UK\\
    \texttt{bpwu1@sheffield.ac.uk}}

\begin{document}
\maketitle
\begin{abstract}
This paper addresses the limitations of the common data annotation and training methods for objective single-label classification tasks. Typically, in such tasks annotators are only asked to provide a single label for each sample and annotator disagreement is discarded when a final hard label is decided through majority voting. We challenge this traditional approach, acknowledging that determining the appropriate label can be difficult due to the ambiguity and lack of context in the data samples. Rather than discarding the information from such ambiguous annotations, our soft label method makes use of them for training. Our findings indicate that additional annotator information, such as confidence, secondary label and disagreement, can be used to effectively generate soft labels. Training classifiers with these soft labels then leads to improved performance and calibration on the hard label test set.
\end{abstract}

\section{Introduction}

Reliable, human-annotated data is crucial for training and evaluation of classification models, with the quality of annotations directly impacting the models' classification performance. Traditionally, in order to ensure high quality annotated data, multiple annotators are asked to judge each individual data instance, and the final `gold standard' \textit{hard label} is determined by majority vote.

However, this hard label approach tends to ignore valuable information from the annotation process, failing to capture the uncertainties and intricacies in real-world data \citep{uma2021learning}. An emerging alternative approach that addresses these limitations is the use of soft labels through techniques such as Knowledge Distillation \citep{hinton2015distilling}, Label Smoothing \citep{szegedy2016rethinking}, Confidence-based Labeling \citep{collins2022eliciting}, and Annotation Aggregation \citep{uma2020case}. These soft label approaches demonstrate potential for improved robustness \cite{peterson2019human}, superior calibration, enhanced performance \cite{fornaciari-etal-2021-beyond} and even enable less than one-shot learning \cite{sucholutsky2021less}. 

This paper's primary focus is on exploring effective ways for improving classification performance using soft labels. Our experimental findings indicate that confidence-based labelling significantly enhances model performance. Nevertheless, the interpretation of confidence scores can also profoundly influence model capability. Given the variability in confidence levels among different annotators \cite{lichtenstein1977those}, aligning these disparate confidence levels emerges as the central research question of this paper.

To address this challenge, we propose a novel method for generating enhanced soft labels by leveraging annotator agreement to align confidence levels. Our contributions include:
\setlist{nolistsep}
    \begin{itemize}[noitemsep]    \item We demonstrate how classification performance can be improved by using soft labels generated from annotator confidence and secondary labels. This presents a solution to the challenge of generating high-quality soft labels with limited annotator resources.
    \item We propose a Bayesian approach to leveraging annotator agreement as a way of aligning individual annotators' confidence scores.

    \item We introduce a novel dataset to facilitate research on the use of soft labels in Natural Language Processing.\footnote{Dataset can be found at: \url{https://github.com/GateNLP/dont-waste-single-annotation}}
\end{itemize}

\section{Related Work}
Current research typically interprets annotator disagreement in two primary ways, either by capturing diverse beliefs among annotators, or by assuming a single ground truth label exists despite disagreement \citep{rottger-etal-2022-two,uma2021learning}. This paper focuses on situations where the latter viewpoint is more applicable. Thus, we focus on traditional ``hard'' evaluation metrics such as F1-score which rely on a gold-label, despite the emergence of alternative, non-aggregated evaluation approaches \citep{basile-etal-2021-need,baan-etal-2022-stop,basile2020s}. This is made possible because we evaluate on high-agreement test sets, where the `true' label is fairly certain.  

Aggregation of annotator disagreement generally falls into two categories: aggregating labels into a one-hot \textit{hard label} \citep{dawid1979maximum,hovy-etal-2013-learning,jamison-gurevych-2015-noise,beigman-beigman-klebanov-2009-learning}, or modeling disagreement as a probability distribution with \textit{soft labels} \citep{sheng2008get, uma2020case,peterson2019human,davani-etal-2022-dealing,rodrigues2018deep,fornaciari-etal-2021-beyond}. 

Similar to \citet{collins2022eliciting}, our study explores how soft labels can be generated from a small pool of annotators, using additional information such as their self-reported confidence. This has benefits over traditional hard/soft label aggregation, which requires extensive annotator resources and/or reliance on potentially unreliable crowd-sourced annotators \citep{snow-etal-2008-cheap,dumitrache-etal-2018-crowdsourcing,poesio-etal-2019-crowdsourced,nie-etal-2020-learn}.

\section{Methodology}

\begin{figure}
\centering
\includegraphics[width=\linewidth]{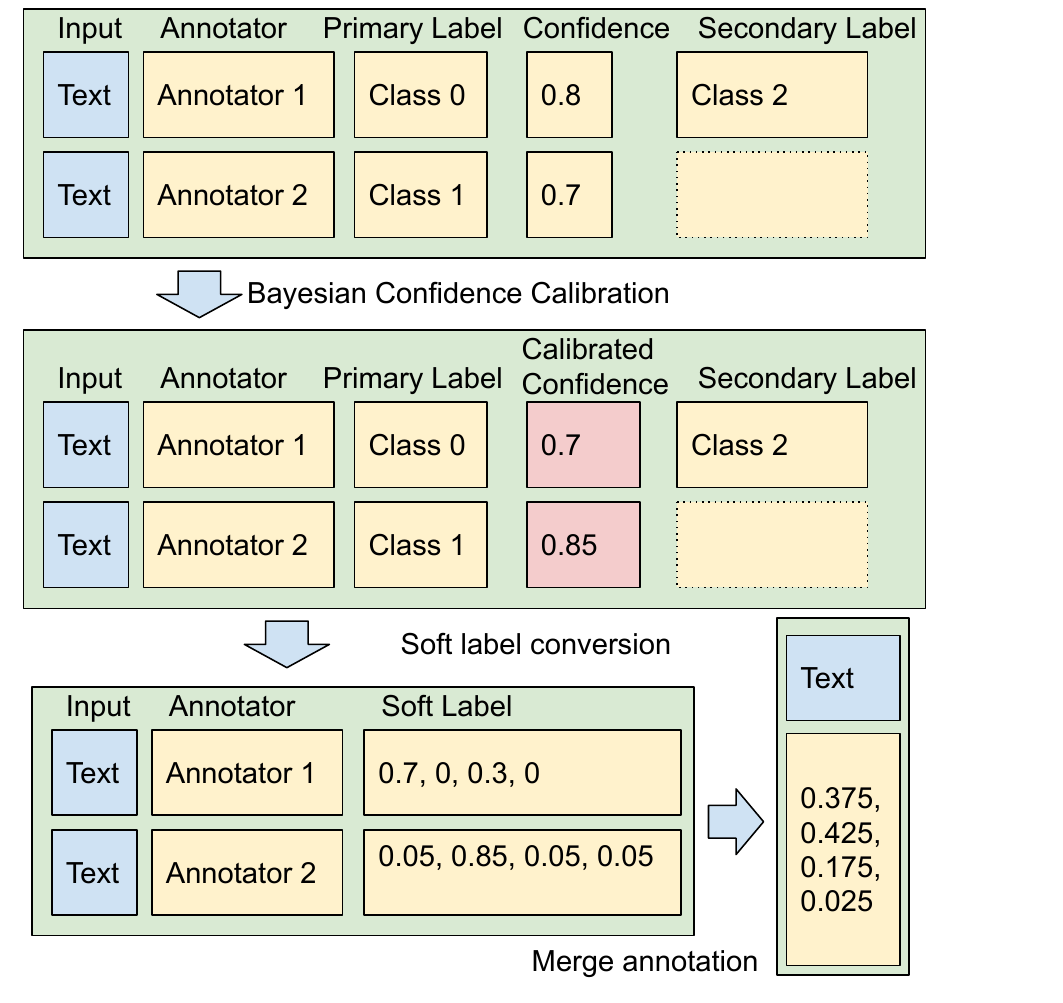}
\vspace{-8mm}
\caption{Soft label conversion pipeline}
\label{fig:pipeline}
\vspace{-6mm}
\end{figure}

In order to generate soft labels, our methodology requires annotators to provide confidence scores. Figure~\ref{fig:pipeline} shows how each annotator provides both a primary class label and a confidence rating that represents their certainty. This ranges from 0 to 1, with 1 representing 100\% confidence.\footnote{In practice, annotators can provide this directly as a percentage or choose from a Likert-style numerical rating (e.g. 1-5) that is then converted to 0-1 scale.} In addition, annotators can also provide an optional `secondary' class label. This represents their selection of the second most probable class. 
Thus, formally, the annotation of the text $x_i$ by an annotator $a_m$ consists of a primary label $l_{im}$, its confidence rating $c_{im}$, and an optional secondary label $l^2_{im}$. We use $y_i$ to denote the text's true label.

Overall, there are three steps to generating soft labels as shown in Figure~\ref{fig:pipeline}:\begin{enumerate}
    \item Annotator confidences are calibrated using our Bayesian method (Section \ref{sec:Bayesian})
    \item Annotations are converted to soft labels (Section \ref{sec:Soft Label Conversion})
    \item Annotator soft labels are merged into a final soft label (Section \ref{sec:Soft Label Conversion})
\end{enumerate}

\subsection{Bayesian Confidence Calibration} \label{sec:Bayesian}

To calibrate confidence levels across annotators, we use annotator agreement as a proxy for reliability. For each annotator, we consider the level of agreement obtained (across all their annotations) when they have expressed a particular confidence score, and use this to re-weight their confidence. The process comprises two steps:

First, we compute the probability of the primary label ($l_{im}$) according to the confidence level ($c_{im}$). This step is agnostic to the identity of the annotator. 
\vspace{-3mm}
\begin{equation}
\vspace{-2mm}
P(\hat{y}_i=l_{im} | c_{im} ) =  \frac{P(c_{im}|l_{im}) P(l_{im})}{P(c_{im})}
\label{eq:prior}
\end{equation} 

\noindent Where $P(l_{im})$ is the prior, $P(c_{im}|l_{im})$ is the likelihood of the confidence score assigned to the primary label and $P(c_{im}) = P(c_{im}|l_{im}) P(l_{im}) + P(c_{im}|\neg l_{im}) P(\neg l_{im})$ is the marginal probability. In this paper, we just make a simple assumption $P(l_{im}) = 1/C$ and $P(\neg l_{im}) = (C-1)/C$, where $C$ is the total number of possible classes. We also assume that $P(c_{im}|y_i) = c_{im}$ and $P(c_{im}|\neg y_i) = 1 - c_{im}$.   

In the second step, using information about agreement, we compute the calibrated probability of the primary label ($l_{im}$) given the specifc annotator ($a_{m}$). 
\vspace{-3mm}
\begin{equation}
\vspace{-2mm}
P(y_i=l_{im} | a_{m} ) =  \frac{P(a_{m}|l_{im}) P(l_{im})}{P(a_{m})}
\label{eq:posterior}
\end{equation} 

\noindent Where our updated prior $P(l_{im}) = P(\hat{y}_i)$ calculated from Equation~\ref{eq:prior}. $P(a_{m}|l_{im})$ is the likelihood that an annotator assigns the label $ l_{im}$ matching the true label, $y_i$, and ${P(a_{m})} = P(a_{m}|y_i) P(\hat{y}_i) + P(a_{i}|\neg y_i) P(\neg \hat{y}_i)$.


We calculate this likelihood based on annotator disagreement: 
\begin{equation}
P(a_{m}|y_i) = \frac{Count(a_{m} \cap \neg a_{m}, a_{m}, l{i})}{Count(a_{m} \cap \neg a_{m}, \neg a_{m}, l{i})}
\end{equation} 

$Count(a_{m} \cap \neg a_{m}, a_{m}, l{i})$ is the number of samples that involves annotator $a_{m}$ and any other annotator(s) $\neg a_{m}$, where both $a_{m}$ and at least one other annotator provided the label $l{i}$. $Count(a_{m} \cap \neg a_{m}, \neg a_{m}, l{i})$ is the number of samples that involves annotator $a_{m}$ and other annotator(s) $\neg a_{m}$, where at least one annotator provided the label $l{i}$.

Similarly, we calculate: 
\begin{equation}
P(a_{m}|\neg y_i) = \frac{Count(a_{m} \cup \neg a_{m}, a_{m},l{i})} {Count(a_{m} \cap \neg a_{m}, \neg a_{m}, \neg l{i})} 
\end{equation}

$Count(a_{m} \cup \neg a_{m}, a_{m},l{i})$ is the number of samples that involves annotator $a_{m}$ and other annotator(s) $\neg a_{m}$ where $a_{m}$ annotated the sample as $l{i}$ but others did not. $Count(a_{m} \cap \neg a_{m}, \neg a_{m}, \neg l{i})$ is the number of samples that involves annotator $a_{m}$ and other annotator $\neg a_{m}$ where $\neg a_{m}$ did not annotate the sample as label $\ l{i}$.


We also employ fallback mechanisms to mitigate cases where annotator agreement cannot be reliably estimated. If the number of counts (< 3 in our experiments) is insufficient to calculate posterior probability, we fall back to using the prior confidence score.







\subsection{Soft Label Conversion \& Merging Annotations} \label{sec:Soft Label Conversion}
Once we have calibrated confidence, $P(y_i | a_{m} )$, we assign this probability to the primary class $l_i$, and assign $1-P(y_i | a_{m} )$ to the secondary class $l^2_i$. Thus, in Figure \ref{fig:pipeline}, Annotator 1's soft label consists of 0.7 for their primary class and 0.3 for their secondary class. If a secondary class label is not provided by the annotator, we just uniformly distribute the confidence level to the other classes. In Annotator 2's case, this means 0.85 for their primary class and 0.05 for the three remaining classes.

Once we have generated soft labels for each annotator, we merge these into a final soft label by taking the mean of each class.

\section{Experiments}
\begin{table*}[ht!]
\centering
\scalebox{0.7}{
\begin{tabular}{l|cccccccc}
\hline
& \multicolumn{3}{c}{\textbf{VaxxHesitancy (5 fold)}} &  \multicolumn{3}{c}{\textbf{CDS (5 fold)}}\\ 
&\textbf{Labels Used} & \textbf{F1 Macro} & \textbf{ECE Calibration} &\textbf{Labels Used} &\textbf{F1 Macro} & \textbf{ECE Calibration}\\ \hline

\multicolumn{1}{c|}{Test-set only}  & & \\\hline
Hard Label & 1 and 2 & 66.9  & 0.262  & 1 & 65.2  & 0.182 \\
Soft Label & 1 and 2 &  \textbf{68.7}  & \textbf{0.192}  & 1 & \textbf{67.7}  & \textbf{0.161} \\ \hline

\multicolumn{1}{c|}{Train+Test-set}   & & \\\hline
Without confidence  & &\\ \hline 
Hard Label & 1 and 2 & 70.2 & 0.237  & 1 & 67.1  & 0.211  \\ 
Dawid Soft Label & 1  & \textbf{71.7}  & \textbf{0.162}  & 1 & 19.6  & -\\ 

Soft Label (primary label) & 1  & 71.0  & 0.171  & 1 & \textbf{68.2}  & \textbf{0.160}  \\ 
Soft Label (primary and secondary) & 1 and 2  & 70.7  & 0.188 & - & - & -\\ \hline

With confidence  & & \\\hline
Hard Label & 1 and 2 & 73.1  & 0.198  & 1 & 69.0  & 0.190 \\
Hard Label (label smoothed 0.1) & 1 and 2 & 68.8  & 0.167  & 1 & 68.8  & \textbf{0.096}  \\

Soft Label (from primary label) & 1 & 73.7  & 0.107  & 1 & \textbf{71.1}  & \textbf{0.096} \\
Soft Label (primary and secondary) & 1 and 2 & \textbf{74.5}  & \textbf{0.106}  & - & - & -\\ \hline

With confidence + bayes calibration & & & \\ \hline

Bayesian Soft Label & 1 and 2 &  \textbf{75.2} & \textbf{0.099} & 1 & 70.4 & 0.118\\
\hline
\hline
\end{tabular}
}
\vspace{-2mm}
\caption{Evaluation results for the CDS and VaxxHesitancy datasets. The 'Labels Used' column indicates whether the hard/soft labels are generated using only the primary label (1), or considering the secondary label as well (2). For the CDS dataset, all labels are generated using only primary since no secondary labels are available.} 
\label{tab:merged}
\vspace{-4mm}
\end{table*}

\subsection{Datasets}
We experiment with two datasets: VaxxHesitancy \citep{mu2023vaxxhesitancy} and the COVID-19 Disinformation Corpus (CDS) \citep{song2021classification}. Both datasets release the confidence scores which annotators provided alongside their class labels (annotators are denoted by a corresponding anonymous ID). 

\subsubsection{COVID-19 Disinformation Corpus (CDS)}

CDS \citep{song2021classification} includes 1,480 debunks of COVID-19-related disinformation from various countries. The debunks are classified into ten topic categories (e.g., public authority, conspiracies and prominent actors). The number of annotators per instance ranges from one to six. Each annotator has provided only one first-choice topic class and their confidence score for each annotated debunk ($ 0\leq c_{im} \leq 9$).

\subsubsection{VaxxHesitancy} 

VaxxHesitancy \citep{mu2023vaxxhesitancy} consists of 3,221 tweets annotated for stance towards the COVID-19 vaccine. Each instance is categorised into pro-vaccine, anti-vaccine, vaccine-hesitant, or irrelevant. The number of annotators per tweet ranges from one to three. Annotators provide a first-choice stance category and a confidence score ($ 1\leq c_{im} \leq 5$). 


\paragraph{VaxxHesitancy Additional Annotation} \label{sec:addilabel}

As our aim is to investigate how additional information provided by annotators could impact classification performance, we also explore the integration of a secondary label for instances where annotators have expressed uncertainty about their primary label choice.

As none of the original datasets had such secondary labels, we undertake an additional round of data annotation, based on the original annotation guidelines. We introduce two new tasks in this data annotation round: 1) For all instances (train + test set) exhibiting low confidence (less than 4), we optionally request that annotators provide a `second stance' label. We guide annotators to propose this if they believe it to be appropriate, even if it wasn't their primary choice. Consequently, we add 569 additional second-choice stances. 2) We assign a third annotator to all instances annotated by two annotators. As a result, we obtain a majority vote for the majority of annotated tweets. This majority vote can be employed for hard label training in subsequent experiments.






\subsubsection{Dataset Split}

We construct the test sets to contain instances where annotators reach agreement on every instance with high confidence scores. 
For VaxxHesitancy, we follow the original train-test split by including instances whose confidence scores are larger than three in the test set. For CDS, the test set has debunks that are labelled by more than one annotator with confidence scores larger than six (on their original 10 point scale). Given the limited size of this subset, data with only one annotation but very high confidence scores is also included in the CDS test set. Summary of the statistics is in the Appendix Table \ref{tab:datasets}.


\subsection{Baselines \& Ablations}

We compare our methods against a variety of hard/soft label aggregation strategies, which make use of annotator confidences/secondary labels to varying degrees.

\noindent{\bf Hard label w/o confidence} We employ majority voting for hard labels. In the absence of consensus, a class category is chosen at random. 

\noindent{\bf Hard label with confidence:} For each $x_i$, we aggregate $a_i$ to estimate a single hard label $\hat{y}_{i}$ by giving different weights to $l_{im}$ according to the annotator confidence $c_{im}$. 


\noindent {\bf Dawid-Skene Soft label:}We utilise an enhanced Dawid-Skene model \citep{passonneau-carpenter-2014-benefits} as an alternative to majority voting, and use the model's confusion matrices to generate soft labels. This model only relies on class labels does not make use of additional information.







\noindent{\bf Label Smoothing Soft Label:}
For each class, we use a mixture of its one-hot hard label vector and the uniform prior distribution over this class \citep{szegedy2016rethinking}.

\noindent{\bf Soft label w/o annotator confidence} We explore generating soft labels using only annotator disagreement. In this approach, we assign 0.7 to the primary stance, 0.3 to the secondary stance label (if available), or evenly distribute the remaining probability among all other classes.



\subsection{Experimental Setup}

We conduct 5-fold cross-validation where each fold contains the entire training set, and 4/5 folds of the test set. We also investigate a second scenario in which we perform 5-fold cross-validation only on the test set. These two scenarios allow us to investigate the performance of soft labels when the level of annotator agreement differs. 

These two scenarios are motivated by the fact that the train-test splits of our datasets contain an uneven distribution of samples: low annotator agreement samples were placed in the train set and high-agreement samples in the test set. This is necessary to enable evaluation against a gold-standard test set. However, for generating soft labels, we want to use a mixture of high-agreement and low-agreement annotations, so we include a portion of the original test set for training.  

We perform experiments using Pytorch \citep{paszke2019pytorch} and the Transformers library from HuggingFace \citep{wolf-etal-2020-transformers}. We fine-tune COVID-Twitter BERT V2,  a BERT large uncased model that has been pre-trained on COVID-19 data \citep{10.3389/frai.2023.1023281}. We fine-tune for 20 epochs with learning rate as 2e-5 (1 epoch warm-up followed by linear decay) and batch size as 8, with AdamW optimizer \citep{loshchilov2017decoupled}. We use cross-entropy loss. The model performance is evaluated with macro-F1 due to the imbalanced datasets, and we use expected calibration error (ECE) \citep{naeini2015obtaining} to measure model calibration.

For both the VaxxHesitancy and CDS datasets, we harmonise the range of confidence scores. In the case of the VaxxHesitancy dataset, this involves converting a confidence score of 5 to 1, 4 to 0.9, and so forth, down to 1 being converted to 0.6. Similarly, for the CDS dataset, a confidence score of 10 is converted to 1, 9 to 0.95, and so on, with 1 also becoming 0.6. \footnote{We manually tested different confidence conversion scales and this conversion yields the best classification performance. See Appendix \ref{tab:covidresults} for an alternative conversion strategy.}


\section{Results}

\textbf{Soft labels improve classification performance across all of our scenarios.}
Table \ref{tab:merged} presents the results on the VaxxHesitancy and CDS datasets. Soft labels surpass hard labels for both datasets, with and without confidence, as well as for the test-only and train + test set scenarios. In the case of test-set only, soft labels achieve 68.7 F1 Macro vs hard label's 66.9 (VaxxHesitancy) and 67.7 vs 65.2 (CDS). As previously mentioned, the test set comprises of only high-agreement samples, so this indicates that soft labels are beneficial for learning even when there is not a lot of disagreement between annotators and they are relatively certain.


\paragraph{Combining confidence scores and secondary labels generates better soft labels.} Using annotators' self-reported confidence scores helps to improve soft labels, as shown by the F1 and calibration improvements between soft labels in the `with' and `without confidence' settings (Table \ref{tab:merged}). Alternative approaches such as Dawid Skene are able to outperform soft labels when confidence scores are not available (71.7 vs 71.0 on VaxxHesitancy). However, once confidence information is introduced, soft labels significantly improves and outperforms alternatives, achieving 73.7 (from 71.0) on VaxxHesitancy and 71.1 (from 68.2 on CDS). 

Furthermore, for the VaxxHesitancy dataset, once secondary label information is included, classification performance is further improved from 73.7 to 74.5. This suggests that more consideration should be taken during the data annotation stage to collect such information, as it can be greatly beneficial for the creation of effective soft labels.


\paragraph{Bayesian calibration outperforms other methods on the VaxxHesitancy dataset.}
By incorporating the full annotation information, i.e., confidence, secondary label, and annotator agreement, our Bayesian soft label method achieves a 75.2 F1 Macro score on the VaxxHesitancy dataset, improving upon 74.5 from the soft label stance 1 and 2. In addition, Bayesian soft label also has the best confidence alignment (ECE) score. 

However, on the CDS dataset, despite outperforming hard labels, our Bayesian method fails to improve upon soft labels. Its adjustments to soft labels results in a fall from 71.1 to 70.4. We believe this is due to the characteristics of the CDS dataset, which has 10 possible classes (vs the 4 of VaxxHesitancy), an increased range of possible confidences (1-9), as well as fewer overall samples. Because there is less annotator overlap,  this greater range of options makes it more difficult to accurately estimate annotator reliability on a per-confidence score level. This reveals a direction in which our Bayes methodcould be improved, as it is currently reliant on sufficient overlap between an individual annotator and their peers.

\section{Conclusion}

We demonstrate the benefits of using soft labels over traditional hard labels in classification tasks. We propose a novel Bayesian method for annotation confidence calibration, and efficiently utilising all available annotation information, outperforming other methods for the VaxxHesitancy dataset. The performance improvements offered by soft labels suggests the importance of collecting additional information from annotators during the annotation process, with annotator confidence being particularly important.

\section*{Acknowledgments}
This work has been co-funded by the UK’s innovation agency (Innovate UK) grant 10039055 (approved under the Horizon Europe Programme as vera.ai, EU grant agreement 101070093).\footnote{\url{https://www.veraai.eu/}}, the European Union under action number 2020-EU-IA-0282 and agreement number INEA/CEF/ICT/A2020/2381686 (EDMO Ireland).\footnote{\url{https://edmohub.ie}}. Ben Wu is supported by an EPSRC Doctoral Training Partnership Grant and Yue Li is supported by a Sheffield–China Scholarships Council PhD Scholarship.
\section*{Limitations}


Our dataset construction introduces secondary labels that are not provided by the same annotators as those who created the original dataset, which may not accurately reflect the choices that the original annotators would have made. 

As discussed in the Results section, the CDS dataset was more challenging due number of class labels and number of samples. This caused the Dawid Skene model to perform poorly. This issue may be alleviated using Laplace Smoothing, but we did not explore this due to time constraints. 


Another important limitation of our Bayesian method is its assumption that an individual annotator's level of agreement with their peers is a good proxy for their reliability. This leaves it vulnerable to situations where there is high agreement between poor annotators.

Even though our soft label method is effective, it is not compared against `traditional' soft labels, which are constructed by aggregating many annotator labels per sample, since this would necessitate the large-scale annotation of the two datasets by many users, which we are trying to avoid as our goal is to reduce the amount of annotators and annotation effort required. 

Finally, we observed high variance across folds during cross-validation. We believe this was due to the small size of test set as well as the presence of ‘hard-to-classify’ samples in certain folds. These were samples where annotators relied on multimodal information to come to a decision (e.g. viewing a video embedded in the tweet). Our model is only provided with text, and so struggles on these samples. 




\section*{Ethics Statement}
Our work has received ethical approval from the Ethics Committee of our university and complies with the research policies of Twitter and follows established protocols for the data annotation process. The human annotators were recruited and trained following our university's ethics protocols, including provision of an information sheet, a consent form, and the ability to withdraw from annotation at any time. For quality and ethics reasons, volunteer annotators were recruited from our university, rather than via Mechanical Turk.

\bibliography{anthology,custom}

\begin{thebibliography}{32}
\expandafter\ifx\csname natexlab\endcsname\relax\def\natexlab#1{#1}\fi

\bibitem[{Baan et~al.(2022)Baan, Aziz, Plank, and
  Fernandez}]{baan-etal-2022-stop}
Joris Baan, Wilker Aziz, Barbara Plank, and Raquel Fernandez. 2022.
\newblock \href {https://aclanthology.org/2022.emnlp-main.124} {Stop measuring
  calibration when humans disagree}.
\newblock In \emph{Proceedings of the 2022 Conference on Empirical Methods in
  Natural Language Processing}, pages 1892--1915, Abu Dhabi, United Arab
  Emirates. Association for Computational Linguistics.

\bibitem[{Basile et~al.(2021)Basile, Fell, Fornaciari, Hovy, Paun, Plank,
  Poesio, and Uma}]{basile-etal-2021-need}
Valerio Basile, Michael Fell, Tommaso Fornaciari, Dirk Hovy, Silviu Paun,
  Barbara Plank, Massimo Poesio, and Alexandra Uma. 2021.
\newblock \href {https://doi.org/10.18653/v1/2021.bppf-1.3} {We need to
  consider disagreement in evaluation}.
\newblock In \emph{Proceedings of the 1st Workshop on Benchmarking: Past,
  Present and Future}, pages 15--21, Online. Association for Computational
  Linguistics.

\bibitem[{Basile et~al.(2020)}]{basile2020s}
Valerio Basile et~al. 2020.
\newblock \href {https://ceur-ws.org/Vol-2776/paper-4.pdf} {It’s the end of
  the gold standard as we know it. on the impact of pre-aggregation on the
  evaluation of highly subjective tasks}.
\newblock In \emph{CEUR WORKSHOP PROCEEDINGS}, volume 2776, pages 31--40.
  CEUR-WS.

\bibitem[{Beigman and
  Beigman~Klebanov(2009)}]{beigman-beigman-klebanov-2009-learning}
Eyal Beigman and Beata Beigman~Klebanov. 2009.
\newblock \href {https://aclanthology.org/P09-1032} {Learning with annotation
  noise}.
\newblock In \emph{Proceedings of the Joint Conference of the 47th Annual
  Meeting of the {ACL} and the 4th International Joint Conference on Natural
  Language Processing of the {AFNLP}}, pages 280--287, Suntec, Singapore.
  Association for Computational Linguistics.

\bibitem[{Collins et~al.(2022)Collins, Bhatt, and
  Weller}]{collins2022eliciting}
Katherine~M Collins, Umang Bhatt, and Adrian Weller. 2022.
\newblock Eliciting and learning with soft labels from every annotator.
\newblock In \emph{Proceedings of the AAAI Conference on Human Computation and
  Crowdsourcing}, volume~10, pages 40--52.

\bibitem[{Davani et~al.(2022)Davani, D{\'\i}az, and
  Prabhakaran}]{davani-etal-2022-dealing}
Aida~Mostafazadeh Davani, Mark D{\'\i}az, and Vinodkumar Prabhakaran. 2022.
\newblock \href {https://doi.org/10.1162/tacl_a_00449} {Dealing with
  disagreements: Looking beyond the majority vote in subjective annotations}.
\newblock \emph{Transactions of the Association for Computational Linguistics},
  10:92--110.

\bibitem[{Dawid and Skene(1979)}]{dawid1979maximum}
Alexander~Philip Dawid and Allan~M Skene. 1979.
\newblock \href {https://www.jstor.org/stable/2346806?origin=crossref} {Maximum
  likelihood estimation of observer error-rates using the em algorithm}.
\newblock \emph{Journal of the Royal Statistical Society: Series C (Applied
  Statistics)}, 28(1):20--28.

\bibitem[{Dumitrache et~al.(2018)Dumitrache, Aroyo, and
  Welty}]{dumitrache-etal-2018-crowdsourcing}
Anca Dumitrache, Lora Aroyo, and Chris Welty. 2018.
\newblock \href {https://doi.org/10.18653/v1/W18-5503} {Crowdsourcing semantic
  label propagation in relation classification}.
\newblock In \emph{Proceedings of the First Workshop on Fact Extraction and
  {VER}ification ({FEVER})}, pages 16--21, Brussels, Belgium. Association for
  Computational Linguistics.

\bibitem[{Fornaciari et~al.(2021)Fornaciari, Uma, Paun, Plank, Hovy, and
  Poesio}]{fornaciari-etal-2021-beyond}
Tommaso Fornaciari, Alexandra Uma, Silviu Paun, Barbara Plank, Dirk Hovy, and
  Massimo Poesio. 2021.
\newblock \href {https://doi.org/10.18653/v1/2021.naacl-main.204} {Beyond black
  {\&} white: Leveraging annotator disagreement via soft-label multi-task
  learning}.
\newblock In \emph{Proceedings of the 2021 Conference of the North American
  Chapter of the Association for Computational Linguistics: Human Language
  Technologies}, pages 2591--2597, Online. Association for Computational
  Linguistics.

\bibitem[{Hinton et~al.(2015)Hinton, Vinyals, and Dean}]{hinton2015distilling}
Geoffrey Hinton, Oriol Vinyals, and Jeff Dean. 2015.
\newblock Distilling the knowledge in a neural network.
\newblock \emph{arXiv preprint arXiv:1503.02531}.

\bibitem[{Hovy et~al.(2013)Hovy, Berg-Kirkpatrick, Vaswani, and
  Hovy}]{hovy-etal-2013-learning}
Dirk Hovy, Taylor Berg-Kirkpatrick, Ashish Vaswani, and Eduard Hovy. 2013.
\newblock \href {https://aclanthology.org/N13-1132} {Learning whom to trust
  with {MACE}}.
\newblock In \emph{Proceedings of the 2013 Conference of the North {A}merican
  Chapter of the Association for Computational Linguistics: Human Language
  Technologies}, pages 1120--1130, Atlanta, Georgia. Association for
  Computational Linguistics.

\bibitem[{Jamison and Gurevych(2015)}]{jamison-gurevych-2015-noise}
Emily Jamison and Iryna Gurevych. 2015.
\newblock \href {https://doi.org/10.18653/v1/D15-1035} {Noise or additional
  information? leveraging crowdsource annotation item agreement for natural
  language tasks.}
\newblock In \emph{Proceedings of the 2015 Conference on Empirical Methods in
  Natural Language Processing}, pages 291--297, Lisbon, Portugal. Association
  for Computational Linguistics.

\bibitem[{Lichtenstein and Fischhoff(1977)}]{lichtenstein1977those}
Sarah Lichtenstein and Baruch Fischhoff. 1977.
\newblock Do those who know more also know more about how much they know?
\newblock \emph{Organizational behavior and human performance}, 20(2):159--183.

\bibitem[{Loshchilov and Hutter(2017)}]{loshchilov2017decoupled}
Ilya Loshchilov and Frank Hutter. 2017.
\newblock \href {https://arxiv.org/abs/1711.05101} {Decoupled weight decay
  regularization}.
\newblock \emph{arXiv preprint arXiv:1711.05101}.

\bibitem[{Mu et~al.(2023)Mu, Jin, Grimshaw, Scarton, Bontcheva, and
  Song}]{mu2023vaxxhesitancy}
Yida Mu, Mali Jin, Charlie Grimshaw, Carolina Scarton, Kalina Bontcheva, and
  Xingyi Song. 2023.
\newblock \href {https://arxiv.org/pdf/2301.06660.pdf} {Vaxxhesitancy: A
  dataset for studying hesitancy towards covid-19 vaccination on twitter}.
\newblock \emph{arXiv preprint arXiv:2301.06660}.

\bibitem[{Müller et~al.(2023)Müller, Salathé, and
  Kummervold}]{10.3389/frai.2023.1023281}
Martin Müller, Marcel Salathé, and Per~E. Kummervold. 2023.
\newblock \href {https://doi.org/10.3389/frai.2023.1023281}
  {Covid-twitter-bert: A natural language processing model to analyse covid-19
  content on twitter}.
\newblock \emph{Frontiers in Artificial Intelligence}, 6.

\bibitem[{Naeini et~al.(2015)Naeini, Cooper, and
  Hauskrecht}]{naeini2015obtaining}
Mahdi~Pakdaman Naeini, Gregory Cooper, and Milos Hauskrecht. 2015.
\newblock Obtaining well calibrated probabilities using bayesian binning.
\newblock In \emph{Proceedings of the AAAI conference on artificial
  intelligence}, volume~29.

\bibitem[{Nie et~al.(2020)Nie, Zhou, and Bansal}]{nie-etal-2020-learn}
Yixin Nie, Xiang Zhou, and Mohit Bansal. 2020.
\newblock \href {https://doi.org/10.18653/v1/2020.emnlp-main.734} {What can we
  learn from collective human opinions on natural language inference data?}
\newblock In \emph{Proceedings of the 2020 Conference on Empirical Methods in
  Natural Language Processing (EMNLP)}, pages 9131--9143, Online. Association
  for Computational Linguistics.

\bibitem[{Passonneau and Carpenter(2014)}]{passonneau-carpenter-2014-benefits}
Rebecca~J. Passonneau and Bob Carpenter. 2014.
\newblock \href {https://doi.org/10.1162/tacl_a_00185} {The benefits of a model
  of annotation}.
\newblock \emph{Transactions of the Association for Computational Linguistics},
  2:311--326.

\bibitem[{Paszke et~al.(2019)Paszke, Gross, Massa, Lerer, Bradbury, Chanan,
  Killeen, Lin, Gimelshein, Antiga et~al.}]{paszke2019pytorch}
Adam Paszke, Sam Gross, Francisco Massa, Adam Lerer, James Bradbury, Gregory
  Chanan, Trevor Killeen, Zeming Lin, Natalia Gimelshein, Luca Antiga, et~al.
  2019.
\newblock \href
  {https://proceedings.neurips.cc/paper_files/paper/2019/file/bdbca288fee7f92f2bfa9f7012727740-Paper.pdf}
  {Pytorch: An imperative style, high-performance deep learning library}.
\newblock \emph{Advances in neural information processing systems}, 32.

\bibitem[{Peterson et~al.(2019)Peterson, Battleday, Griffiths, and
  Russakovsky}]{peterson2019human}
Joshua~C Peterson, Ruairidh~M Battleday, Thomas~L Griffiths, and Olga
  Russakovsky. 2019.
\newblock \href
  {https://openaccess.thecvf.com/content_ICCV_2019/papers/Peterson_Human_Uncertainty_Makes_Classification_More_Robust_ICCV_2019_paper.pdf}
  {Human uncertainty makes classification more robust}.
\newblock In \emph{Proceedings of the IEEE/CVF International Conference on
  Computer Vision}, pages 9617--9626.

\bibitem[{Poesio et~al.(2019)Poesio, Chamberlain, Paun, Yu, Uma, and
  Kruschwitz}]{poesio-etal-2019-crowdsourced}
Massimo Poesio, Jon Chamberlain, Silviu Paun, Juntao Yu, Alexandra Uma, and Udo
  Kruschwitz. 2019.
\newblock \href {https://doi.org/10.18653/v1/N19-1176} {A crowdsourced corpus
  of multiple judgments and disagreement on anaphoric interpretation}.
\newblock In \emph{Proceedings of the 2019 Conference of the North {A}merican
  Chapter of the Association for Computational Linguistics: Human Language
  Technologies, Volume 1 (Long and Short Papers)}, pages 1778--1789,
  Minneapolis, Minnesota. Association for Computational Linguistics.

\bibitem[{Rodrigues and Pereira(2018)}]{rodrigues2018deep}
Filipe Rodrigues and Francisco Pereira. 2018.
\newblock Deep learning from crowds.
\newblock In \emph{Proceedings of the AAAI conference on artificial
  intelligence}, volume~32.

\bibitem[{Rottger et~al.(2022)Rottger, Vidgen, Hovy, and
  Pierrehumbert}]{rottger-etal-2022-two}
Paul Rottger, Bertie Vidgen, Dirk Hovy, and Janet Pierrehumbert. 2022.
\newblock \href {https://doi.org/10.18653/v1/2022.naacl-main.13} {Two
  contrasting data annotation paradigms for subjective {NLP} tasks}.
\newblock In \emph{Proceedings of the 2022 Conference of the North American
  Chapter of the Association for Computational Linguistics: Human Language
  Technologies}, pages 175--190, Seattle, United States. Association for
  Computational Linguistics.

\bibitem[{Sheng et~al.(2008)Sheng, Provost, and Ipeirotis}]{sheng2008get}
Victor~S Sheng, Foster Provost, and Panagiotis~G Ipeirotis. 2008.
\newblock \href
  {https://dl.acm.org/doi/abs/10.1145/1401890.1401965?casa_token=5BSJGKjj3ssAAAAA:cSEfgisNW7Sf1JBlXGy2hKT-2k-8k-gnlBGLfEpInHDRtnnDS0RyUBCxLYy9szw1RqjnWRvhSzU}
  {Get another label? improving data quality and data mining using multiple,
  noisy labelers}.
\newblock In \emph{Proceedings of the 14th ACM SIGKDD international conference
  on Knowledge discovery and data mining}, pages 614--622.

\bibitem[{Snow et~al.(2008)Snow, O{'}Connor, Jurafsky, and
  Ng}]{snow-etal-2008-cheap}
Rion Snow, Brendan O{'}Connor, Daniel Jurafsky, and Andrew Ng. 2008.
\newblock \href {https://aclanthology.org/D08-1027} {Cheap and fast {--} but is
  it good? evaluating non-expert annotations for natural language tasks}.
\newblock In \emph{Proceedings of the 2008 Conference on Empirical Methods in
  Natural Language Processing}, pages 254--263, Honolulu, Hawaii. Association
  for Computational Linguistics.

\bibitem[{Song et~al.(2021)Song, Petrak, Jiang, Singh, Maynard, and
  Bontcheva}]{song2021classification}
Xingyi Song, Johann Petrak, Ye~Jiang, Iknoor Singh, Diana Maynard, and Kalina
  Bontcheva. 2021.
\newblock \href
  {https://journals.plos.org/plosone/article?id=10.1371/journal.pone.0247086}
  {Classification aware neural topic model for covid-19 disinformation
  categorisation}.
\newblock \emph{PloS one}, 16(2):e0247086.

\bibitem[{Sucholutsky and Schonlau(2021)}]{sucholutsky2021less}
Ilia Sucholutsky and Matthias Schonlau. 2021.
\newblock Less than one'-shot learning: Learning n classes from m< n samples.
\newblock In \emph{Proceedings of the AAAI Conference on Artificial
  Intelligence}, volume~35, pages 9739--9746.

\bibitem[{Szegedy et~al.(2016)Szegedy, Vanhoucke, Ioffe, Shlens, and
  Wojna}]{szegedy2016rethinking}
Christian Szegedy, Vincent Vanhoucke, Sergey Ioffe, Jon Shlens, and Zbigniew
  Wojna. 2016.
\newblock \href {https://ieeexplore.ieee.org/document/7780677} {Rethinking the
  inception architecture for computer vision}.
\newblock In \emph{Proceedings of the IEEE conference on computer vision and
  pattern recognition}, pages 2818--2826.

\bibitem[{Uma et~al.(2020)Uma, Fornaciari, Hovy, Paun, Plank, and
  Poesio}]{uma2020case}
Alexandra Uma, Tommaso Fornaciari, Dirk Hovy, Silviu Paun, Barbara Plank, and
  Massimo Poesio. 2020.
\newblock \href {https://ojs.aaai.org/index.php/HCOMP/article/view/7478} {A
  case for soft loss functions}.
\newblock In \emph{Proceedings of the AAAI Conference on Human Computation and
  Crowdsourcing}, volume~8, pages 173--177.

\bibitem[{Uma et~al.(2021)Uma, Fornaciari, Hovy, Paun, Plank, and
  Poesio}]{uma2021learning}
Alexandra~N Uma, Tommaso Fornaciari, Dirk Hovy, Silviu Paun, Barbara Plank, and
  Massimo Poesio. 2021.
\newblock \href {https://www.jair.org/index.php/jair/article/view/12752}
  {Learning from disagreement: A survey}.
\newblock \emph{Journal of Artificial Intelligence Research}, 72:1385--1470.

\bibitem[{Wolf et~al.(2020)Wolf, Debut, Sanh, Chaumond, Delangue, Moi, Cistac,
  Rault, Louf, Funtowicz, Davison, Shleifer, von Platen, Ma, Jernite, Plu, Xu,
  Le~Scao, Gugger, Drame, Lhoest, and Rush}]{wolf-etal-2020-transformers}
Thomas Wolf, Lysandre Debut, Victor Sanh, Julien Chaumond, Clement Delangue,
  Anthony Moi, Pierric Cistac, Tim Rault, Remi Louf, Morgan Funtowicz, Joe
  Davison, Sam Shleifer, Patrick von Platen, Clara Ma, Yacine Jernite, Julien
  Plu, Canwen Xu, Teven Le~Scao, Sylvain Gugger, Mariama Drame, Quentin Lhoest,
  and Alexander Rush. 2020.
\newblock \href {https://doi.org/10.18653/v1/2020.emnlp-demos.6} {Transformers:
  State-of-the-art natural language processing}.
\newblock In \emph{Proceedings of the 2020 Conference on Empirical Methods in
  Natural Language Processing: System Demonstrations}, pages 38--45, Online.
  Association for Computational Linguistics.

\end{thebibliography}
\bibliographystyle{acl_natbib}

\appendix

\section{Appendix}
\label{sec:appendix}

Table \ref{tab:datasets} shows information about the composition of the two datasets used in our experiments.

\begin{table}[h!]
\centering
\scalebox{0.7}{
\begin{tabular}{l|cccc}
\hline
& \multicolumn{2}{c}{\textbf{VaxxHesitancy}} & \multicolumn{2}{c}{\textbf{CDS}} \\ 
&\textbf{Train} & \textbf{Test} & \textbf{Train} & \textbf{Test}\\ \hline
Number of items & 2,790 & 431 & 965 & 515\\
Avg annotations per item & 1.25 & 2 & 1.38 & 1.58\\
Avg confidence score & 4.16 & 4.63 & 6.73 & 8.53\\

\hline

\hline
\end{tabular}
}
\vspace{-2mm}
\caption{Summary of the datasets}
\vspace{-2mm}
\label{tab:datasets}
\end{table}

Table \ref{tab:covidresults} shows the effect of changing the annotator confidence conversion scale from the one presented in the main section of this paper (9: 1.0 ... 1:0.6) to an alternative one (9: 1.0, 8: 0.9 ... 1: 0.1). By comparing between the two columns, we can see that this change leads to a drop of ~1 F1 point for the resulting soft labels. This highlights the importance of selecting a good initial conversion scale.  

\begin{table*}[!t]
\centering
\scalebox{0.8}{
\begin{tabular}{l|cccc}
\hline
& \multicolumn{2}{c}{\textbf{Covid Misinfo (5 fold)}} \\ 
& &  \multicolumn{1}{c}{\textbf{1-0.1 conversion}} & \multicolumn{1}{c}{\textbf{1-0.6 conversion}} \\ 
& \textbf{Uses confidence} & \textbf{F1 Macro} & \textbf{F1 Macro}\\ \hline

Experiments on test-set only & & & \\\hline
Hard Label & yes & 65.18 & - (same as left) \\
Soft Label & yes & \textbf{66.51} & \textbf{67.71} \\ \hline

Experiments on train + test-set & & & \\\hline
Hard Label (majority vote, ties broken by confidence) & yes & 68.53 & 69.02 \\
Soft Label & yes & \textbf{70.59} & \textbf{71.11} \\ \hline

Experiments with no confidence & & & \\ \hline 
Hard Label & no & 67.13 & - (same as left) \\
Soft Label & no & \textbf{68.15} & - (same as left) \\ 
Dawid Soft Label & no & 19.58 & - (same as left) \\ \hline

Experiments with confidence & & & \\\hline
Hard Label & yes & 68.53 & 69.02 \\
Hard Label (label smoothed 0.1) & yes & 68.69 & 68.75  \\
Hard Label (label smoothed 0.3) & yes & 68.86 & 68.44 \\
Soft Label & yes & \textbf{70.59} & \textbf{71.11} \\ \hline

Experiments with bayesian calibration & & & \\\hline
Soft label with Bayesian & yes & 69.75 & 70.40 \\
\hline

\hline
\end{tabular}
}
\caption{CDS Results with different confidence conversions.}
\label{tab:covidresults}
\end{table*}

\end{document}